\def\year{2022}\relax
\documentclass[letterpaper]{article}
\usepackage{aaai22}
\usepackage{times}
\usepackage{helvet}
\usepackage{courier}
\usepackage[hyphens]{url}
\usepackage{graphicx}
\usepackage{natbib}
\usepackage{caption}
\usepackage{bbm}
\usepackage{float}
\usepackage{stfloats}
\usepackage{graphicx}
\usepackage{amsmath}
\usepackage{amssymb}
\usepackage{subfiles}
\usepackage{subfigure}
\usepackage{algorithm}
\usepackage{algpseudocode}

\newtheorem{definition}{Definition}

\title{GraphMixup: Improving Class-Imbalanced Node Classification on \\ Graphs by Self-supervised Context Prediction}

\author{
Lirong Wu {$^{1,2,3}$}, Haitao Lin{$^{1,2}$}, Zhangyang Gao{$^{1,2,3}$}, Cheng Tan{$^{1,2}$}, Stan.Z.Li{$^{1,2,^\dagger}$}\\
}
\affiliations{
$^1$ AI Lab, School of Engineering, Westlake University, Hangzhou 310024, Zhejiang Province, China \\ 
$^2$ Institute of Advanced Technology, Westlake Institute for Advanced Study, Hangzhou 310024, Zhejiang Province, China \\
$^3$ Zhejiang University, Hangzhou 310058, Zhejiang Province, China \\
\{wulirong, linhaitao, gaozhangyang, tancheng, stan.zq.li\}@westlake.edu.cn
}

\begin{document}

\maketitle

\begin{abstract}
Recent years have witnessed great success in handling node classification tasks with Graph Neural Networks (GNNs). However, most existing GNNs are based on the assumption that node samples for different classes are balanced, while for many real-world graphs, there exists the problem of class imbalance, i.e., some classes may have much fewer samples than others. In this case, directly training a GNN classifier with raw data would under-represent samples from those minority classes and result in sub-optimal performance. This paper presents GraphMixup, a novel mixup-based framework for improving class-imbalanced node classification on graphs. However, directly performing mixup in the input space or embedding space may produce out-of-domain samples due to the extreme sparsity of minority classes; hence we construct semantic relation spaces that allows the \emph{Feature Mixup} to be performed at the semantic level. Moreover, we apply two context-based self-supervised techniques to capture both local and global information in the graph structure and then propose \emph{Edge Mixup} specifically for graph data. Finally, we develop a \emph{Reinforcement Mixup} mechanism to adaptively determine how many samples are to be generated by mixup for those minority classes. Extensive experiments on three real-world datasets show that GraphMixup yields truly encouraging results for class-imbalanced node classification tasks.
\end{abstract}

\vspace{-1.5em}
\section{Introduction}
Recently, the emerging Graph Neural Networks (GNNs) have demonstrated their powerful capability to handle the task of semi-supervised node classification: inferring unknown node labels by using the graph structure and node features with partially known node labels. Despite all these successes, existing works are mainly based on the assumption that node samples for different classes are roughly balanced. However, in many real-world applications, there exists the serious class-imbalanced problem, i.e., some classes may have significantly fewer samples for training than other classes. For example, the majority of users in a transaction fraud network are benign users, while only a small portion of them are bots. Similarly, topic classification for citation networks also suffers from this problem, as the papers for some topics may be scarce, comparing to those on-trend topics.

The class-imbalanced problems have been well studied in the image domain, and data-level algorithms can be summarized into two groups: down-sampling and over-sampling \cite{more2016survey}. The down-sampling methods sample a representative sample set from the majority class to make its size close to the minority class, but this inevitably entails a loss of information. In contrast, the over-sampling methods aim to generate new samples for minority classes, which have been found to be more effective and stable. However, directly applying existing over-sampling strategies to graph data may lead to sub-optimal results due to the non-Euclidean property of graphs. Three key problems for mitigating the class-imbalanced problem on graphs by over-sampling are: \emph{(1) How to generate new nodes and their features for minority classes? (2) How to capture the connections between the generated node and the existing nodes in the graph? (3) How to determine the upsampling scale for each minority class?} 

Mixup \cite{zhang2017mixup,verma2019manifold} is an effective method to solve \textit{Problem (1)}, which performs feature interpolation for minority classes to generate new samples. However, most existing mixup methods are performed either in the input space or embedding space, which may generate out-of-domain samples, especially for those minority classes due to their extreme sparsity. To alleviate this problem, disentangled semantic spaces are constructed in this paper to allow the \emph{Feature Mixup} to be performed at the semantic level. To solve \textit{Problem (2)}, GraphSMOTE \cite{zhao2021graphsmote} proposes to train an edge generator through the task of adjacency matrix reconstruction and then applies it to predict the existence of edges between generated nodes and existing nodes. However, MSE-based matrix reconstruction completely ignores local and global structural information, making the edge generator overemphasize the connections between nodes with similar features while neglecting the long-range dependencies between nodes. Therefore, we design two context-based self-supervised tasks to consider both local and global information in the graph structure. Finally, unlike heuristic estimation for \textit{Problem (3)}, we develop a reinforcement mixup mechanism to adaptively determine the upsampling scale for each minority class.

Our main contributions are summarized as follows:

\begin{itemize}
    \item Disentangled semantic spaces are constructed to perform \emph{Semantic Feature Mixup} at the semantic level.
    \item Propose \emph{Contextual Edge Mixup} specifically for graphs and apply two context-based self-supervised techniques to consider both local and global structure information.
    \item Develop a reinforcement mixup mechanism instead of heuristic hyperparameters to adaptively determine the upsampling ratio for each minority class.
    \item Extensive experiments on three real-world datasets show that GraphMixup outperforms other leading methods covering the full spectrum of low-to-high imbalance ratios.
\end{itemize}

\vspace{-1em}
\section{Related Work}
\textbf{Class-Imbalanced Problem.}
The class-imbalanced problem is common in real-world scenarios and has become a popular research topic \cite{johnson2019survey,rout2018handling}. The mainstream algorithms can be divided into two categories: algorithm-level and data-level. The algorithm-level methods \cite{ling2008cost,zhou2005training,parambath2014optimizing} seek to directly increase the importance of minority classes with suitable penalty functions. Instead, the data-level methods usually adjust class sizes through down-sampling or over-sampling. In this paper, we mainly focus on solving the class-imbalanced problem for graph data with \emph{oversampling-like algorithms}. The vanilla over-sampling is replicating existing samples, which reduces the class imbalance but can lead to over-fitting as no extra information is introduced. SMOTE \cite{chawla2002smote} solves this problem by generating new samples by feature interpolation between samples of minority classes and their nearest neighbors, and many of its variants \cite{han2005borderline,bunkhumpornpat2009safe} have been proposed with promising results. However, most previous efforts focused on the image domain, and few attempts have been made on class-imbalanced problems for non-Euclidean graph data. GraphSMOTE \cite{zhao2021graphsmote} is the first work to consider the problem of node-class imbalance on graphs, but their contribution is only to extend SMOTE to graph settings without making full use of the semantic feature information and local/global structural information embedded in graph data.

\noindent \textbf{Disentanglement Learning.}
The disentanglement aims to decompose an entity, such as a feature vector, into several independent components to better capture semantic information. Most recent works are based on the autoencoder architecture, where the latent features generated by the encoder are constrained to be independent in each dimension. The works of DisenGCN \cite{ma2019disentangled} and IPGDN\cite{liu2020independence}, as pioneering attempts, achieve node-level disentanglement through neighbor routines that divide the neighbors of a node into several mutually exclusive parts. FactorGCN \cite{yang2020factorizable}, on the other hand, performs relation disentanglement by taking into account global topological semantics. The semantic disentanglement method proposed in this paper is similar to FactorGNN in that the disentangled semantic features are learned for each node by considering higher-order semantic relations between nodes.

\noindent \textbf{Graph Self-Supervised Learning (SSL).}
The primary goal of Graph SSL is to learn transferable prior knowledge from abundant unlabeled data with well-designed pretext tasks and then generalize the learned knowledge to downstream tasks. The existing graph SSL methods can be divided into three categories: contrastive, generative, and predictive \cite{wu2021self}. The contrastive methods contrast the views generated from different augmentation by mutual information maximization. Instead, the generative methods focus on the (intra-data) information embedded in the graph, generally based on pretext tasks such as reconstruction. Moreover, the predictive methods generally self-generate labels by some simple statistical analysis or expert knowledge and then perform prediction-based tasks based on self-generated labels. In this paper, we mainly focus on context-based self-supervised prediction since it takes full account of the contextual information in the graph structure, both local and global, allowing us to better capture connections between generated nodes and existing nodes.

\begin{figure*}[!htbp]
	\begin{center}
		\includegraphics[width=1.0\linewidth]{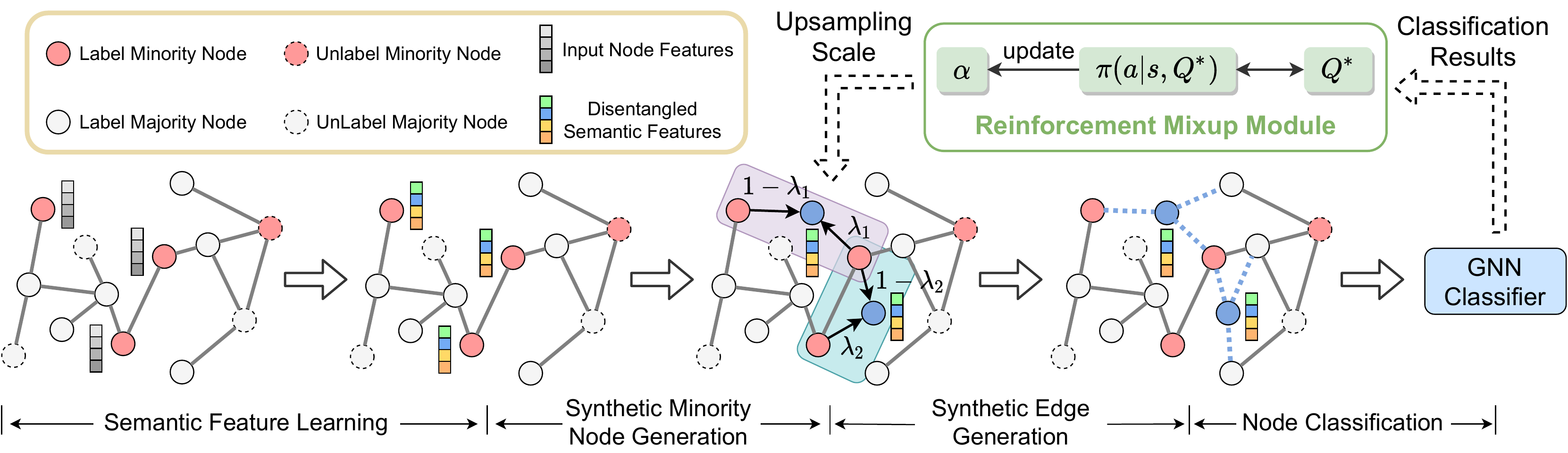}
	\end{center}
	\caption{Illustration of the GraphMixp framework, which consists of the following steps: (1) learning disentangled semantic features by constructing semantic relation spaces; (2) generate synthetic minority nodes by semantic-level feature mixup; (3) generate synthetic edges by performing edge mixup with an edge predictor trained on two well-designed context-based self-supervised tasks; (4) Classify using a GNN classifier and feed the results back to the RL agent to update the upsampling scale.}
	\label{fig:1}
\end{figure*}

\vspace{-0.5em}
\section{Methodology}
\vspace{-0.5em}
\subsection{Problem Statement}
Given an input graph $\mathcal{G}=(\mathcal{V}, \mathcal{E})$, where $\mathcal{V}$ is the set of $N$ nodes with features $\mathcal{X}=\left(\mathbf{x}_{1}, \mathbf{x}_{2}, \cdots, \mathbf{x}_{N}\right)\in \mathbb{R}^{N \times F}$ and $\mathcal{E} \subseteq \mathcal{V} \times \mathcal{V}$ is the set of edges. Each node $v \in \mathcal{V}$ is associated with an features vector $x_{v} \in \mathcal{X}$, and each edge $e_{u, v} \in \mathcal{E}$ denotes a connection between node $u$ and node $v$. The graph structure can also be represented by an adjacency matrix $\mathbf{A} \in[0,1]^{N \times N}$ with $A_{u,v}=1$ if $e_{u,v}\in\mathcal{E}$ and $A_{u,v}=0$ if $e_{u,v} \notin \mathcal{E}$. We first define the concepts and notions about node class-imbalance ratio: 

\vspace{-0.5em}
\begin{definition}
Suppose there are $m$ classes of nodes $\mathcal{C}=\left\{C_{1}, \ldots, C_{m}\right\}$ in the graph $\mathcal{G}$, where $|C_{i}|$ is the number of samples belong to $i$-th class. Class-Imbalance Ratio $h=\frac{\min _{i}\left(\left|C_{i}\right|\right)}{\max _{i}\left(\left|C_{i}\right|\right)}$ is the ratio of the size of the largest majority class to the smallest minority class in the graph $\mathcal{G}$.
\end{definition}
\vspace{-0.5em}

Node classification is a typical node-level task where only a subset of node $\mathcal{V}_L$ with corresponding features $\mathcal{X}_L$ and labels $\mathcal{Y}_L$ are known, and we denote the labeled set as $\mathcal{D}_L=(\mathcal{V}_L,\mathcal{X}_L,\mathcal{Y}_L)$ and unlabeled set as $\mathcal{D}_U=(\mathcal{V}_U,\mathcal{X}_U,\mathcal{Y}_U)$. The purpose of GraphMixup is to perform feature, label and edge mixups for minority classes $\mathcal{C}_S\subseteq\mathcal{C}$ to generate a synthetic set $\mathcal{D}_S=(\mathcal{V}_S,\mathcal{X}_S,\mathcal{Y}_S)$ and its corresponding edge set $\mathcal{E}_S=\{e_{v',u}|v'\in\mathcal{V}_S,u\in\mathcal{V}\}$. Then the synthesized set $\mathcal{D}_S$ is moved into the labeled set $\mathcal{D}_L$ to obtain a updated labeled set $\mathcal{D}_N=\mathcal{D}_{L} \bigcup \mathcal{D}_{S}$. Similarly, we can obtain an updated edge set $\mathcal{E}_N=\mathcal{E}_{L} \bigcup \mathcal{E}_{S}$ as swell as its corresponding adjacency matrix $\mathbf{A}_N$, where $\mathbf{A}_N[:N,:N]=\mathbf{A}$. Let $\Phi: \mathcal{V} \rightarrow \mathcal{Y}$ be a graph network trained on labeled data $\mathcal{D}_N$ so that it can be used to infer the labels $\mathcal{Y}_U$ of unlabeled data.

In this paper, we present the details of the proposed GraphMixup framework, with an overview shown in Fig.~1. The main idea of GraphMixup is to perform feature mixup to generate synthetic minority nodes in disentangled semantic spaces by a \emph{Semantic Feature Mixup} module. Next, two context-based self-supervised pretext tasks are applied to train a \emph{Contextual Edge Mixup} module that captures both local and global connections between generated nodes and existing for synthetic edge generation. Finally, we detail the \emph{Reinforcement Mixup} mechanism, which can adaptively determine the number of samples to be generated (upsampling scale) by mixup for minority classes.

\subsection{Semantic Feature Mixup}
One effective way to generate minority nodes is to apply feature mixup directly in the input space or embedding space. However, this may lead to sub-optimal results since samples of minority classes are usually quite scarce, resulting in a sparse distribution of samples in the input and embedding space, which in turn produces out-of-domain samples during the interpolation process. Therefore, we consider higher-order relations between samples to learn disentangled semantic features through a \emph{semantic feature extractor}, and thus perform semantic-level feature mixup. To this end, we first construct several semantic relation spaces, represented by semantic relation graphs. Then, we perform feature aggregation and transformation in each semantic space separately, and finally merge the semantic features from each space into a concatenated disentangled semantic feature.

\noindent \textbf{Semantic Relation Learning.}
Specifically, we first transform the input nodes to a low-dimensional space, done by multiplying the features of nodes with a parameter matrix $\mathbf{W}_h \in \mathbb{R}^{F_h \times F}$, that is $\mathbf{h}^{\prime}_i=\mathbf{W}_h x_i$. The transformed features are then used to generate a semantic relation graph with respect to semantic relation $k (1 \leq k \leq K)$ as follows
\begin{equation}
G_{k,i,j} = \sigma\big(\Omega_k(\mathbf{h}^{\prime}_i, \mathbf{h}^{\prime}_j)\big)
\end{equation}

\noindent where $\sigma=\tanh(\cdot)$ is an activation function, and $\Omega_k(\cdot)$ is a function that takes the concated features of node $i$ and node $j$ as input and takes the form of an one-layer MLP in our implementation. However, without any other constraints, some of the generated relation graphs may contain similar structures. More importantly, it is not easy to directly maximize the gap between various semantic relation graphs due to the non-Euclidean property of graph structure. Therefore, we first derive a graph descriptor $\mathbf{d}_{k}$ for each relation graph $G_k$,
\begin{equation}
\mathbf{d}_{k}=f\Big(\operatorname{Readout}\big(\mathcal{A}(G_{k}, \mathbf{H}^{\prime})\big)\Big)
\end{equation}

\noindent where $\mathcal{A}(\cdot)$ is a two-layer graph autoencoder \cite{kipf2016variational} which takes $\mathbf{H}^{\prime}=\{\mathbf{h}^{\prime}_1,\mathbf{h}^{\prime}_2,\cdots,\mathbf{h}^{\prime}_N\}$ as inputs, and generates new features for each node, $\operatorname{Readout}(\cdot)$ performs global average pooling for all nodes, and $f(\cdot)$ is a fully connected layer. Note that all semantic relation graphs share the same node features $\mathbf{H}^{\prime}$, making sure that the information discovered by the feature extractor comes only from the differences between graph structures rather than node features. The loss used to train the extractor is defined as 
\begin{equation}
\mathcal{L}_{dis}=\sum_{i=1}^{K-1} \sum_{j=i+1}^{K}\frac{\mathbf{d}_i\cdot \mathbf{d}_j^T}{\|\mathbf{d}_i\|\|\mathbf{d}_j\|}
\end{equation}

\noindent \textbf{Disentangled Semantic Feature Learning.} 
Once the semantic relation learning is completed, the disentangled semantic-specific features can be learned by taking the weighted sum of its neighbors for $l$-th ($1\leq l \leq L$) layer,
\begin{equation}
\mathbf{h}_{i, k}^{(l)}=\sigma\Big(\sum_{j \in N_{i, k}} G_{k, i, j} \mathbf{W}^{(l, k)} \mathbf{h}_{j}^{(l-1)}\Big)
\end{equation}

\vspace{-0.5em}
\noindent where $\mathbf{h}_j^{(0)}=\mathbf{x}_j$ and $\mathbf{h}_{i, k}^{(l)}$ represents the semantic feature of node $i$ with respect to relation $k$ in $l$-th layer. In the semantic relation graph $G_k$, $\mathcal{N}_{i, k}$ is the neighbours of node $i$, $G_{k, i, j}$ is the weighting coefficient from node $i$ to node $j$, and $\mathbf{W}^{(l, k)} \in \mathbb{R}^{F_h \times F_h}$ is a parameter matrix. Finally, the learned features from different semantic relation space can be merged to produce disentangled node features, as follows
\begin{equation}
\mathbf{h}_{i}^{(l)}=\|_{k=1}^{K} \mathbf{h}_{i, k}^{(l)}
\end{equation}

\noindent \textbf{Synthetic Minority Node Generation.}
After obtaining the disentangled semantic features for each node by semantic feature extractor, we can perform semantic-level feature mixup to generate new samples for minority classes. Specifically, we perform interpolation on sample $v$ from one target minority class with its nearest neighbor $n n(v)$, as follows
\begin{equation}
\begin{split}
\mathbf{h}_{v^{\prime}}^{(L)}&=(1-\delta) \cdot \mathbf{h}_{v}^{(L)}+\delta \cdot \mathbf{h}_{n n(v)}^{(L)} \\ 
n n(v)&=\underset{u\in\{\mathcal{V}/v\}, y_u=y_v}{\operatorname{argmin}}\left\|\mathbf{h}_{u}^{(L)}-\mathbf{h}_{v}^{(L)}\right\|
\end{split}
\label{equ:6}
\end{equation}

\noindent where $\delta$ is a random variable, following uniform distribution in the range [0, 1]. Since node $v$ and $nn(v)$ belong to the same class and are very close to each other, the generated node $v'$ should also belong to the same class. In this way, the \emph{label mixup} can be simplified to directly assign the same label as the source node $v$ to the newly synthesized node $v'$.

\vspace{-0.5em}
\subsection{Contextual Edge Mixup}
Now we have generated synthetic node $\mathcal{V}_S$, node feature $\mathcal{X}_S$, and label $\mathcal{Y}_S$ by means of feature mixup and label mixup described above. However, these new synthetic nodes are still isolated from the raw graph $\mathcal{G}$ and do not have any links with the nodes in the raw node set $\mathcal{V}$. Therefore, we introduce \emph{edge mixup} to capture the connections between generated nodes and existing nodes. To this end, we design an edge prediction that is trained on the raw node set $\mathcal{V}$ and edge set $\mathcal{E}$ and then used to predict relation connectivity between generated nodes in the set $\mathcal{V}_S$ and existing nodes in the set $\mathcal{V}$. Specifically, we implement the edge predictor as:
\begin{equation}
\widehat{\mathbf{A}}_{v, u}=\sigma\left(\mathbf{z}_{v} \cdot  \mathbf{z}_{u}^T\right); \mathbf{z}_u = \overline{\mathbf{W}} \mathbf{h}_u^{(L)}, \mathbf{z}_v = \overline{\mathbf{W}} \mathbf{h}_v^{(L)}
\end{equation}

\noindent where $\widehat{\mathbf{A}}_{v, u}$ refers to the predicted relation connectivity between node $v$ and $u$, and $\overline{\mathbf{W}} \in \mathbb{R}^{F_h \times F_h}$ is the parameter matrix. The loss function for training the edge predictor is
\begin{equation}
\mathcal{L}_{rec}=\|\widehat{\mathbf{A}}-\mathbf{A}\|_{F}^{2}
\end{equation}

Since the above MSE-based matrix reconstruction only considers the connectivity between nodes based on feature similarity, it may ignore important information of the graph structure, so we employ two additional context-based self-supervised prediction tasks to capture both local and global structural information for a better edge predictor. 

\noindent \textbf{Context-based Self-supervised Prediction.}
The first pretext task \textit{Local-Path Prediction} is to predicte the shortest path length between different node pairs. To prevent very noisy ultra-long pairwise distances from dominating the optimization, we truncate the shortest path longer than 4, which also \emph{forces the model to focus on the local structure}. Specifically, it first randomly samples a certain amount of node pairs $\mathcal{S}$ from all node pairs $\{(v,u)|v,u\in\mathcal{V}\}$ and calculates the pairwise node shortest path length $d_{v,u}=d(v,u)$ for each node pair $(v,u)\in\mathcal{S}$. Furthermore, it groups the shortest path lengths into four categories: $C_{v,u}=0, C_{v,u}=1, C_{v,u}=2$, and $C_{v,u}=3$ corresponding to $d_{v,u}=1, d_{v,u}=2, d_{v,u}=3$, and $d_{v,u}\geq3$, respectively. The learning objective is then formulated as a multi-class classification problem, as follows
\begin{equation}
\mathcal{L}_{local}=\frac{1}{|\mathcal{S}|}\sum_{(v, u) \in \mathcal{S}} \ell\Big(f_\omega^{(1)}\big(|\mathbf{z}_v-\mathbf{z}_u|\big), C_{v,u}\Big) 
\end{equation}

\vspace{-1em}
\noindent where $\ell(\cdot)$ denotes the cross-entropy loss and $f_\omega^{(1)}(\cdot)$ linearly maps the input to a 4-dimension value.

The second pretext task \textit{Global-Path Prediction} pre-obtains a set of clusters from raw node set $\mathcal{V}$ and then guides the model to \emph{preserve global topology information} by predicting the shortest path from each node to the anchor nodes associated with cluster centers. Specifically, it first partitions the graph into $T$ clusters $\{M_1,M_2,\cdots,M_T\}$ by applying unsupervised graph partition algorithm \cite{karypis1998fast}. Inside each cluster $M_t$ ($1\leq t \leq T$), the node with the highest degree is taken as corresponding cluster center, denoted as $m_t$ . Then it calculates the distance $\mathbf{l}_i \in \mathbb{R}^T$ from node $v_i$ to cluster centers $\{m_k\}_{k=1}^T$. The learning objective is then formulated as a regression problem, defined as
\begin{equation}
\begin{aligned}
\mathcal{L}_{global}= \frac{1}{|\mathcal{V}|}\sum_{v_{i} \in \mathcal{V}} \left\|f_\omega^{(2)}\left(\mathbf{z}_i\right)-\mathbf{l}_i\right\|^2
\end{aligned}
\end{equation}

\vspace{-1em}
\noindent where $f_\omega^{(2)}(\cdot)$ linearly maps the input to $K$-dimension values. The total loss to train the edge predictor is defined as
\begin{equation}
\mathcal{L}_{edge} = \mathcal{L}_{rec} + \mathcal{L}_{local} + \mathcal{L}_{global}
\end{equation}

Context-based self-supervised methods have been proposed in other work \cite{jin2020self,peng2020self} as auxiliary tasks to help feature extraction. However, we apply self-supervised tasks for learning a better edge predictor rather than for learning transferable knowledge on unlabeled data. More importantly, the two self-supervised tasks described above capture both local and global information in the graph structure, which makes them \emph{more beneficial for edge prediction as opposed to the task of feature extraction}.

\noindent \textbf{Synthetic Edge Generation.}
With the learned edge predictor, we can perform \emph{Edge Mixup} in two different ways. The first scheme is to directly use \emph{continuous edges}, that is
\begin{equation}
    \mathbf{A}_N[v',u]=\widehat{\mathbf{A}}_{v', u}
    \label{equ:12}
\end{equation}
\vspace{-1em}

\noindent where $v'\in\mathcal{V}_S$ and $u\in\mathcal{V}$. The second scheme is to obtain the \emph{binary edges} by setting a threshold value, as follows
\vspace{-0.5em}
\begin{equation}
\mathbf{A}_N[v',u]=\left\{\begin{array}{ll}
1, & \text { if } \widehat{\mathbf{A}}_{v', u}>\eta \\
0, & \text { otherwise }
\end{array}\right.
\label{equ:13}
\end{equation}
\vspace{-0.9em}

\noindent The above two strategies are both implemented in this paper denoted as $\text{GraphMixup}_C$ and $\text{GraphMixup}_B$ respectively, and their performance are compared in the experiment part.

\vspace{-0.8em}
\subsection{Reinforcement Mixup Mechanism}
\vspace{-0.2em}
The upsampling scale, i.e., the number of synthetic samples to be generated by mixup, is important for model performance. A too large scale may introduce redundant and noisy information, while a too small scale is not efficient enough to alleviate the class-imbalanced problem. Therefore, instead of setting the upsampling scale $\alpha$ as a fixed hyperparameter for all minority classes and then estimating it heuristically, we use a novel reinforcement learning algorithm that adaptively updates the upsampling scale for each minority class. We model the updating process as a Markov Decision Process (MDP) \cite{white1989markov}. Formally, the state, action, transition, reward, and termination are defined as:

\noindent $\bullet$ $\textit{\textbf{State.}}$ For minority class set $\mathcal{C}_S$, the state $s_e$ at epoch $e$ is represented by the number of new samples for each minority class, that is $s_e=\{|C_i|\cdot\alpha_i\}_{C_i\in\mathcal{C}_S}$, where $\alpha_i=\alpha_i^{init}+\kappa_i$.

\noindent $\bullet$ $\textit{\textbf{Action.}}$ RL agent updates $\{\kappa_i\}_{C_i\in\mathcal{C}_S}$ by taking action $a_e$ based on reward. We define the action $a_e$ as add or minus a fixed value $\Delta\kappa$ from $\{\kappa_i\}_{C_i\in\mathcal{C}_S}$ at each epoch $e$.

\noindent $\bullet$ $\textit{\textbf{Transition.}}$ We generate $|C_i|\cdot\alpha_i$ new synthetic nodes as defined in Eq.~(\ref{equ:6}) for each minority class in the next epoch.

\noindent $\bullet$ $\textit{\textbf{Reward.}}$ Due to the black-box nature of GNN, it is hard to sense its state and cumulative reward. So we define a discrete reward function $\text{reward}\left(s_{e}, a_{e}\right)$ for each action $a_e$ at state $s_e$ directly based on the classification results, as follows
\begin{small}
\begin{equation}
reward\left(s_{e}, a_{e}\right)=\left\{\begin{array}{cl}
+1, & \text { if } cla_{e}>cla_{e-1} \\
0, & \text { if } cla_{e}=cla_{e-1} \\
-1, & \text { if } cla_{e}<cla_{e-1}
\end{array}\right.
\label{equ:14}
\end{equation}
\end{small}

\noindent where $cla_e$ is the macro-F1 score at epoch $e$. Eq.~(\ref{equ:14}) indicates that if the macro-F1 with action $a_e$ is higher than the previous epoch, the reward for $a_e$ is positive, and vice versa.

\noindent $\bullet$ $\textit{\textbf{Termination.}}$ If the change of $\{\kappa_i\}_{C_i\in\mathcal{C}_S}$ among twenty consecutive epochs is no more than $\Delta\kappa$, the RL algorithm will stop, and $\{\kappa_i\}_{C_i\in\mathcal{C}_S}$ will remain fixed during the next training process.  The terminal condition is formulated as:
\begin{equation}
\operatorname{Range}\left(\left\{\kappa_i^{e-20}, \cdots, \kappa_i^{e}\right\}\right) \leq T_\kappa,\quad C_i \in \mathcal{C}_S
\end{equation}

The $Q$-learning \cite{watkins1992q} is applied to learn the above MDP. $Q$-learning is an off-policy reinforcement learning algorithm that seeks to find best actions given the current state. It fits the Bellman optimality equation,
\begin{small}
\begin{equation}
Q^{*}\left(s_{e}, a_{e}\right)=\text { reward }\left(s_{e}, a_{e}\right)+\gamma \underset{a^{\prime}}{\arg \max } Q^{*}\left(s_{e+1}, a^{\prime}\right)
\end{equation}
\end{small}

\vspace{-1.5em}
\noindent where $\gamma\in[0,1]$ is a discount factor of future reward. We adopt a $\varepsilon$-greedy policy with an explore probability $\varepsilon$:
\begin{small}
\begin{equation}
\pi\left(a_{e} \mid s_{e} ; Q^{*}\right)=\left\{\begin{array}{cc}
\text { random action } & \text { w.p. } \varepsilon \\
\underset{a_e}{\arg \max } Q^{*}\left(s_{e}, a\right) & \text{otherwise}
\end{array}\right.
\end{equation}
\end{small}

\vspace{-1em}
\noindent This means that the RL agent explores new states by selecting an action at random with probability $\varepsilon$ instead of only selecting actions based on the max future reward. The RL agent and other modules can be trained jointly in an end-to-end manner. The results in the experiment part verify the effectiveness of the reinforcement mixup mechanism.

\vspace{-0.8em}
\subsection{Optimization Objective \& Training Strategy}
\vspace{-0.2em}
Let $\mathbf{P}$ be a new embedding matrix by concatenating the semantic embedding $\mathbf{H}^{(L)}$ of real nodes $\mathcal{V}$ with the semantic embedding $\mathbf{H}_S^{(L)}$ of the synthetics nodes $\mathcal{V}_S$. Then we can obatain label prediction for node $v$ with a \emph{node classifier},
\vspace{-0.5em}
\begin{equation}
\begin{split}
\mathbf{h}_{v}^{(L+1)} &= \sigma\left(\widetilde{\mathbf{W}}^{(1)} \cdot \operatorname{CONCAT}\big(\mathbf{h}_{v}^{(L)}, \mathbf{P}\cdot \widehat{\mathbf{A}}[:, v]\big)\right) \\
\widehat{\mathbf{y}}_v &= softmax(\widetilde{\mathbf{W}}^{(2)} \cdot \mathbf{h}_{v}^{(L+1)})
\end{split}
\end{equation}

\vspace{-0.5em}
\noindent where $\widetilde{\mathbf{W}}^{(1)}\in\mathbb{R}^{F_h \times F_h}$ and $\widetilde{\mathbf{W}}^{(2)}\in\mathbb{R}^{m \times F_h}$ are parameter matrices. The above node classifier is optimized using cross-entropy loss on the updated labeled set $\mathcal{V}_N=\mathcal{V}\bigcup\mathcal{V}_s$ as:
\vspace{-0.5em}
\begin{equation}
\mathcal{L}_{\text {node }}=\sum_{v \in \mathcal{V}_{N}} \sum_{c}\left(\mathbbm{1}\left(\mathbf{y}_v  =c\right) \cdot \log \left(\widehat{\mathbf{y}}_{v}[c]\right)\right.
\end{equation}

\vspace{-0.5em}
As the model performance is dependent on the quality of embedding space and generated edges, to make training phrase more stable, we adopt a two-stage training paradigm. Let $\theta$, $\gamma$, $\phi$ be the parameters for semantic feature extractor, edge predictor, and node classifier respectively. Firstly, the semantic feature extractor and edge predictor are pre-trained with loss $\mathcal{L}_{dis}$ and $\mathcal{L}_{edge}$, then the pre-trained parameters $\theta_{init}$ and $\gamma_{init}$ are used as the initialization. At the fine-tuning stage, the pre-trained encoder $\theta_{init}(\cdot)$ with a node classifier is trained under the supervision of $\mathcal{L}_{node}$. The learning objective is defined as
\vspace{-0.5em}
\begin{equation}
\theta^{*}, \phi^{*}=\arg \min _{(\theta, \phi)} \mathcal{L}_{node}(\theta, \gamma, \phi)
\end{equation}

\vspace{-1em}
\noindent with initialization $\theta_{init}, \gamma_{init}=\arg \min _{(\theta, \gamma)} \mathcal{L}_{dis}(\theta) + \beta\mathcal{L}_{edge}(\gamma)$, where $\beta$ is the weight to balance these two losses. Since $\mathcal{L}_{dis}$ and $\mathcal{L}_{edge}$ are roughly on the same order of magnitude, without loss of generality we set $\beta$ to 1.0 by default (hyperparametric search for $\beta$ may yield better results, but this is not the focus of this paper). The pseudo code of the proposed GraphMixup is summarized in Algorithm \ref{algo:1}.

\vspace{-1em}
\begin{algorithm}[H]
\footnotesize
	\caption{Algorithm for the proposed GraphMixup}
	\label{algo:1}
	\begin{algorithmic}[1]
		\Require Feature Matrix: $\mathbf{X}$; Adjacency Matrix: $\mathbf{A}$.
		
		\Ensure Predicted Labels.

		\State Randomly initialize the semantic feature extractor, edge predictor and node classifier; Initialize upsampling scale $\alpha_i^{init}=\frac{N}{m|C_i|}$ and $\kappa_i=0$ for minority class $C_i\in\mathcal{C}_S$; 
		\State Train the feature extractor and edge predictor until convergence, based on $L_{dis}$ and $L_{edge}$ defined in Eq.~3 and Eq.~11.
		
		\While{Not Converged} 
		    \State \textit{\# Feature Mixup}
		    \State Obtain disentangled features $\mathbf{H}^{(L)}$ by Eq.~4 and Eq.~5;
		    \For{class $i$ in minority classes set $\mathcal{C}_S$}
		        \State Calculate upsampling scale $\alpha_i = \alpha_i^{init}+\kappa_i$
    		    \For{$j$ $\in$ \{0, 1, $\cdots$, $|C_i|*\alpha_i$\}}
    		        \State Generate new samples for class $i$ by Eq.~6;
    		    \EndFor
		    \EndFor
		    \State \textit{\# Edge Mixup}
		    \State Generate new adjacency matrix $\mathbf{A}_N$ by Eq.~12 or Eq.~13;
		    \State Train feature extractor and classifier with $L_{node}$ by Eq.~19;
		    \State \textit{\# RL process}
		    \If Eq.~15 is False
		        \State $reward \left(s_{e}, a_{e}\right) \leftarrow$ Eq.~14;
		        \State $a_e \leftarrow$ Eq.~17;
		        \State $\kappa_i \leftarrow a_e\cdot\Delta\kappa$ for $C_i\in\mathcal{C}_S$;
		        
		    \EndIf
		\EndWhile
		
		\State \textbf{return} Predicted labels $\mathcal{Y}_U$ for unlabeled nodes $\mathcal{V}_U$.
	\end{algorithmic}
\end{algorithm}

\begin{table*}[ht]
\begin{center}
\setlength{\abovecaptionskip}{0.0em}
\setlength{\belowcaptionskip}{-2em}
\caption{Performance comparison of different methods for class-imbalanced node classification.}
\label{tab:1}
\resizebox{\textwidth}{!}{
\begin{tabular}{l|ccc|ccc|ccc}
\hline
 & \multicolumn{3}{c|}{Cora} & \multicolumn{3}{c|}{BlogCatlog} & \multicolumn{3}{c}{Wiki-CS} \\ \hline
Methods & Acc & AUC-ROC & Macro$\pm$F1 & Acc & AUC-ROC & Macro$\pm$F1 & Acc & AUC-ROC & Macro$\pm$F1 \\ \hline
Origin & 0.718$\pm$0.002 & 0.919$\pm$0.002 & 0.715$\pm$0.003 & 0.208$\pm$0.005 & 0.583$\pm$0.004 & 0.067$\pm$0.002 & 0.767$\pm$0.001 & 0.940$\pm$0.002 & 0.735$\pm$0.001 \\
Over-Sampling & 0.731$\pm$0.007 & 0.927$\pm$0.006 & 0.728$\pm$0.008 & 0.202$\pm$0.004 & 0.592$\pm$0.003 & 0.072$\pm$0.003 & 0.779$\pm$0.002 & 0.948$\pm$0.002 & 0.744$\pm$0.002 \\
Re-weight & 0.728$\pm$0.009 & 0.925$\pm$0.005 & 0.724$\pm$0.006 & 0.204$\pm$0.005 & 0.785$\pm$0.004 & 0.069$\pm$0.002 & 0.761$\pm$0.002 & 0.939$\pm$0.002 & 0.738$\pm$0.002 \\
SMOTE & 0.732$\pm$0.010 & 0.925$\pm$0.007 & 0.729$\pm$0.005 & 0.206$\pm$0.004 & 0.795$\pm$0.003 & 0.073$\pm$0.001 & 0.780$\pm$0.004 & 0.945$\pm$0.003 & 0.745$\pm$0.003 \\
Embed-SMOTE & 0.722$\pm$0.006 & 0.918$\pm$0.003 & 0.721$\pm$0.004 & 0.202$\pm$0.006 & 0.781$\pm$0.004 & 0.070$\pm$0.003 & 0.750$\pm$0.005 & 0.943$\pm$0.003 & 0.721$\pm$0.004 \\
GraphSMOTE & 0.742$\pm$0.003 & 0.930$\pm$0.002 & 0.739$\pm$0.002 & 0.247$\pm$0.004 & 0.644$\pm$0.005 & 0.123$\pm$0.002 & 0.785$\pm$0.003 & 0.955$\pm$0.004 & 0.752$\pm$0.003 \\ \hline
$\text{GraphMixup}_B$ & 0.761$\pm$0.001 & 0.934$\pm$0.002 & 0.758$\pm$0.002 & 0.255$\pm$0.003 & 0.663$\pm$0.003 & 0.126$\pm$0.002 & 0.792$\pm$0.002 & 0.958$\pm$0.002 & 0.764$\pm$0.002 \\
$\text{GraphMixup}_C$ & \textbf{0.775$\pm$0.003} & \textbf{0.942$\pm$0.002} & \textbf{0.773$\pm$0.001} & \textbf{0.268$\pm$0.003} & \textbf{0.673$\pm$0.001} & \textbf{0.132$\pm$0.002} & \textbf{0.804$\pm$0.002} & \textbf{0.964$\pm$0.003} & \textbf{0.775$\pm$0.001} \\ \hline
\end{tabular}}
\end{center}
\vspace{-1em}
\end{table*}

\vspace{-2em}
\section{Experiments}
In this section, we show the effectiveness of the proposed GraphMixup on three real-world datasets and provide extensive ablation studies and analysis on its various components. The experiments aim to answer the following five questions:

\noindent $\bullet$ $\textit{\textbf{Q1.}}$ How does GraphMixup perform in class-imbalance node classification on various real-world datasets?

\noindent $\bullet$ $\textit{\textbf{Q2.}}$ Is GraphMixup robust to different imbalance ratios?

\noindent $\bullet$ $\textit{\textbf{Q3.}}$ How does semantic feature extractor (bottleneck encoder) influence the performance of GraphMixup?

\noindent $\bullet$ $\textit{\textbf{Q4.}}$ How do the two context-based self-supervised prediction tasks influence the performance of GraphMixup?

\noindent $\bullet$ $\textit{\textbf{Q5.}}$ How does the reinforcement mixup mechanism work? What happens if the upsampling scale is fixed?

\vspace{-0.5em}
\subsection{Experimental setups}
\noindent \textbf{Datasets.}
The experiments are conducted on three widely used datasets, namely BlogCatalog \cite{tang2009relational}, Wiki-CS \cite{mernyei2020wiki}, and Cora \cite{sen2008collective} datasets. The first one is BlogCatalog dataset, where 14 classes with fewer than 100 samples are taken as minority classes. The second one is Wiki-CS dataset, where we consider classes with fewer than the average samples per class as minority classes. Finally, on the Cora dataset, we randomly selected three classes as minority classes and the rest as majority classes. All majority classes have a training set of 20 samples. For each minority class, the number is 20$\times im\_ratio$ with $im\_ratio$ being 0.5 by default, and we have varied $im\_ratio$ to evaluate the performance of GraphMixup under different imbalanced ratios in the following. 

\noindent \textbf{Baselines.}
To demonstrate the power of GraphMixup to handle class-imbalance problems, we compare it with six baselines: (1) \emph{Origin}: original implementation without additional tricks; (2) \emph{Over-Sampling}: repeat samples directly from minority classes; (3) \emph{Re-weight}: assign higher loss weights to samples from minority classes \cite{yuan2012sampling+}; (4) \emph{SMOTE}: generate synthetic samples by interpolating in the input space, and the edges of newly generated nodes are set to be the same as the source nodes; (5) \emph{Embed-SMOTE}: an extension of SMOTE by interpolating in the embedding space \cite{ando2017deep}; (6) \emph{GraphSMOTE}: an extension of Embed-SMOTE by linking generated nodes to existing nodes through a well-trained edge generator. Basing on strategies for setting edges, two varients of GraphMixup are tested: (7) $\emph{GraphMixup}_B$: the generated edges are set to binary values by thresholding as Eq.~(\ref{equ:13}); (8) $\emph{GraphMixup}_C$: the generated edges are set as continuous values as Eq.~(\ref{equ:12}).

\noindent \textbf{Evaluation Metrics.}
Following existing works in evaluating imbalanced classification, three evaluation metrics are adopted in this paper: Accuracy(\textit{Acc}), AUC-ROC, and Macro-F1. \textit{Acc} is calculated on all test samples at once and thus may underestimate those minority classes. In contrast, both AUC-ROC and Macro-F1 are calculated for each class separately and then non-weighted average over them, thus better reflecting the performance on minority classes.

\noindent \textbf{Hyperparameters.}
The following hyperparameters are set for all datasets: Adam optimizer with learning rate $lr$ = 0.001 and weight decay $decay$ = 5e-4; Maximum Epoch $E$ = 4000; Layer number $L$ = 1 with hidden dimension $d_{F}$ = 32; Semantic Relation $K$ = 4; Loss weights $\alpha$ = 1.0; Threshold $\eta$ = 0.5. In the reinforcement mixup module, we set $\gamma$ = 1, $\varepsilon$ = 0.9, $\Delta \kappa$ = 0.05. Besides, the initial $\kappa_i^{init}$ is set class-wise: $\frac{N}{m|C_i|}$ for minority class $C_i\in\mathcal{C}_S$ on each dataset. Each set of experiments is run 5 times with different random seeds, and the average results are reported as performance metrics.

\vspace{-0.5em}
\subsection{Class-Imbalanced Classification (\textit{Q1})}
To evaluate the effectiveness of GraphMixup in class-imbalanced node classification tasks, we compare it with the other six baselines on three datasets. Table.~\ref{tab:1} shows that the improvements brought by GraphMixup are much larger than directly applying other over-sampling algorithms. For example, compared with GraphSMOTE, $\emph{GraphMixup}_C$ shows an improvement of 3.3\% in \textit{Acc} score and 3.4\% in Macro-F1 score. Moreover, both two variants of GraphSMOTE show significant improvements for imbalanced node classification, compared to almost all baselines on all datasets. Notably, we find that $\emph{GraphMixup}_C$ exhibits slightly better performance than $\emph{GraphMixup}_B$, which implies the advantage of soft continuous edges over thresholded binary edges.

\vspace{-0.5em}
\subsection{Influence of Imbalance Ratio (\textit{Q2})}
The performance under different imbalance ratios is reported in Table.~\ref{tab:2} to evaluate their robustness. Experiments are conducted in the Cora dataset by varying class imbalance ratio $im\_ratio$ as $\{0.1, 0.2, 0.3, 0.4, 0.5, 0.6\}$. The ROC-AUC scores in Table.~\ref{tab:2} show that: (1) GraphMixup generalizes well to different imbalance ratios and achieves the best performance across all settings. (2) The improvement of GraphMixup is more significant when the imbalance ratio is more extreme. For example, when the imbalance ratio is 0.1, $\text{GraphMixup}_C$ outperforms SMOTE by 6.4\%, and the gap reduces 1.5\% when the imbalance ratio reaches 0.6.

\begin{table}[ht]
\begin{center}
\setlength{\abovecaptionskip}{0em}
\setlength{\belowcaptionskip}{-0.5em}
\caption{Performance under different imbalance ratios.}
\label{tab:2}
\resizebox{\columnwidth}{!}{
\begin{tabular}{l|cccccc}
\hline
 & \multicolumn{6}{c}{Class-Imbalanced Ratio} \\ \hline
Methods & 0.1 & 0.2 & 0.3 & 0.4 & 0.5 & 0.6 \\ \hline
Origin & 0.843 & 0.890 & 0.907 & 0.913 & 0.919 & 0.920 \\
Over-Sampling & 0.830 & 0.898 & 0.917 & 0.922 & 0.927 & 0.929 \\
Re-weight & 0.869 & 0.906 & 0.921 & 0.923 & 0.925 & 0.928 \\
SMOTE & 0.839 & 0.897 & 0.917 & 0.924 & 0.925 & 0.929 \\
Embed-SMOTE & 0.870 & 0.897 & 0.906 & 0.912 & 0.918 & 0.925 \\
GraphSMOTE & 0.887 & 0.912 & 0.923 & 0.927 & 0.930 & 0.932 \\ \hline
$\text{GraphMixup}_B$ & 0.898 & 0.915 & 0.923 & 0.932 & 0.934 & 0.935 \\
$\text{GraphMixup}_C$ & \textbf{0.903} & \textbf{0.919} & \textbf{0.931} & \textbf{0.935} & \textbf{0.942} & \textbf{0.944} \\ \hline
\end{tabular}}
\end{center}
\vspace{-2em}
\end{table}

\subsection{Influence of Bottleneck Encoder (\textit{Q3})}
\begin{figure}[!htbp]
	\begin{center}
		\includegraphics[width=1.0\linewidth]{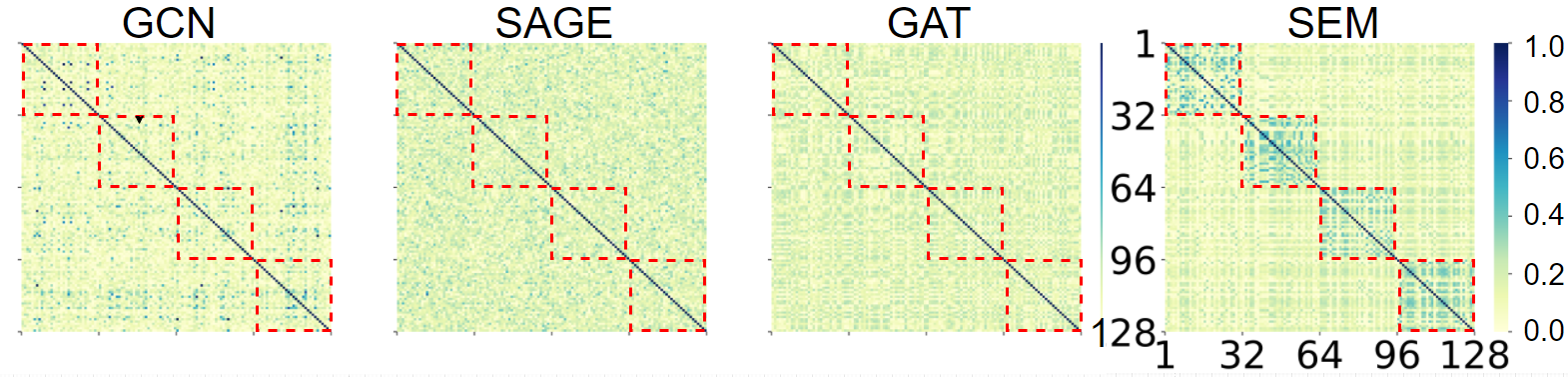}
	\end{center}
	\vspace{-1em}
	\caption{Feature correlation analysis on the Cora dataset.}
	\label{fig:2}
	\vspace{-1em}
\end{figure}

To analyze the effectiveness of the \emph{Semantic Feature Extractor (SEM)} and the applicability of GraphMixup to different bottleneck encoders, we apply three other common encoders: GCN \cite{kipf2016semi}, SAGE \cite{hamilton2017inductive}, and GAT \cite{velivckovic2017graph}. Due to space limitations, only the performance of the AUC-ROC scores on the Cora dataset is reported. Table.~\ref{tab:3} shows that GraphSMOTE works well with all four bottleneck encoders, achieving the best performance. Moreover, results with SEM as the bottleneck encoder are slightly better than the other three across all methods, indicating the benefits of constructing semantic relation spaces, extracting semantic features, and performing semantic-level mixup. Furthermore, Fig.~\ref{fig:2} shows the correlation analysis of 128-dimensional latent features with $K = 4$ semantic relations obtained from four different bottleneck encoders. We find that only the correlation map of SEM exhibits four clear diagonal blocks, which demonstrates its excellent capability to extract highly independent disentangled semantic features. 

\begin{table}[ht]
\begin{center}
\setlength{\abovecaptionskip}{0em}
\setlength{\belowcaptionskip}{-1em}
\caption{Performance with different bottleneck encoders.}
\label{tab:3}
\footnotesize
\begin{tabular}{l|cccc}
\hline
 & \multicolumn{4}{c}{Bottleneck Encoder} \\ \hline
Methods & GCN & SAGE & GAT & SEM \\ \hline
Origin & 0.909 & 0.897 & 0.912 & \textit{0.919} \\
Over-Sampling & 0.916 & 0.907 & 0.923 & \textit{0.927} \\
Re-weight & 0.917 & 0.904 & 0.919 & \textit{0.925} \\
SMOTE & 0.917 & 0.907 & 0.919 & \textit{0.925} \\
Embed-SMOTE & 0.914 & 0.906 & 0.916 & \textit{0.918} \\
GraphSMOTE & 0.920 & 0.914 & 0.923 & \textit{0.930} \\ \hline
$\text{GraphMixup}_B$ & 0.924 & 0.916 & 0.926 & \textit{0.934} \\
$\text{GraphMixup}_C$ & \textbf{0.926} & \textbf{0.919} & \textbf{0.932} & \textit{\textbf{0.942}} \\ \hline
\end{tabular}
\end{center}
\vspace{-0.5em}
\end{table}

\vspace{-0.5em}
\subsection{RL Process Analysis (\textit{Q4})}
\begin{figure}[!htbp]
	\begin{center}
		\subfigure[Performance under different (fixed) upsampling scale.]{\includegraphics[width=0.9\linewidth]{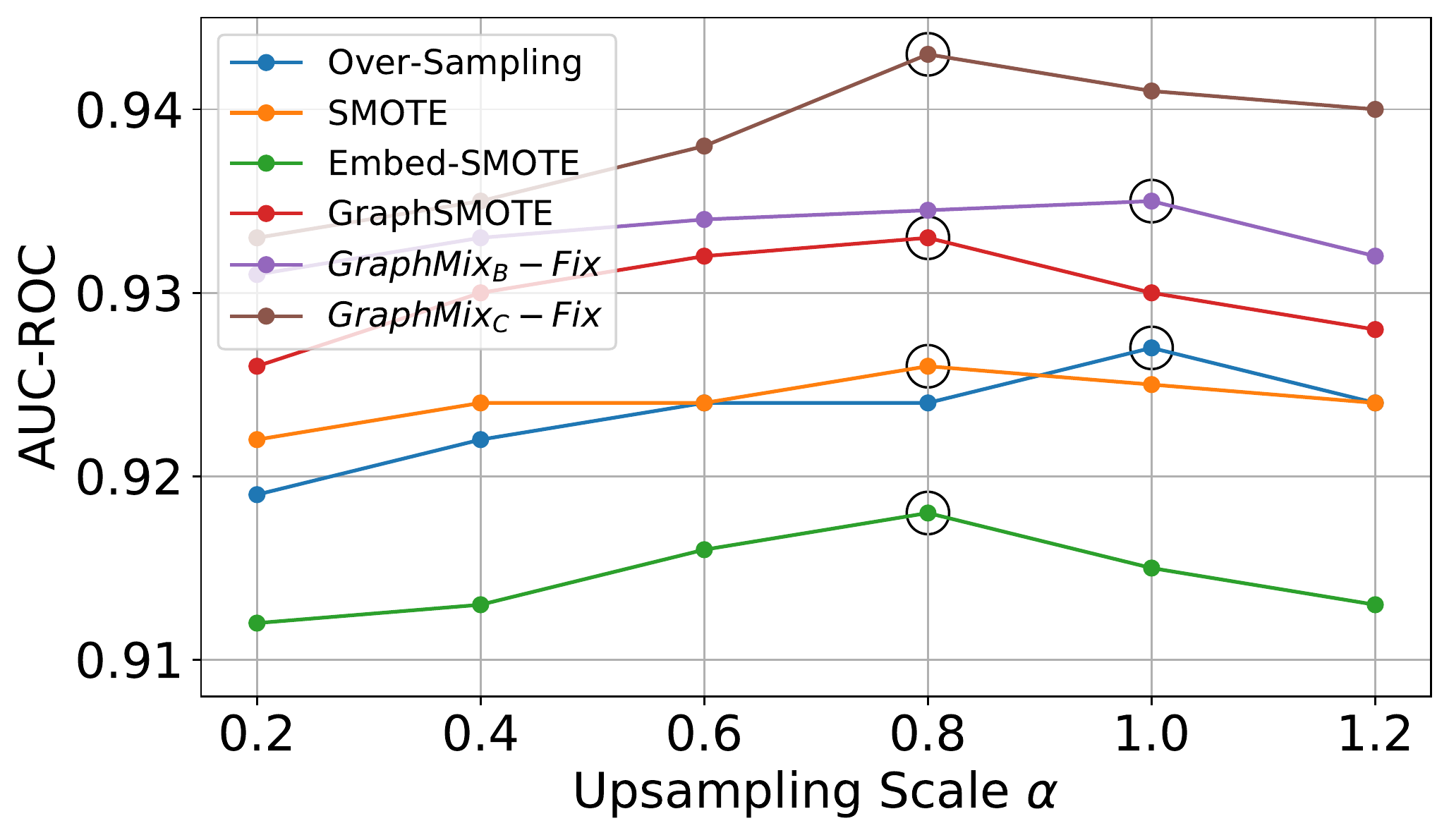}\label{fig:3a}}
		\subfigure[Updating process of the cumulative change in $\kappa$.]{\includegraphics[width=0.9\linewidth]{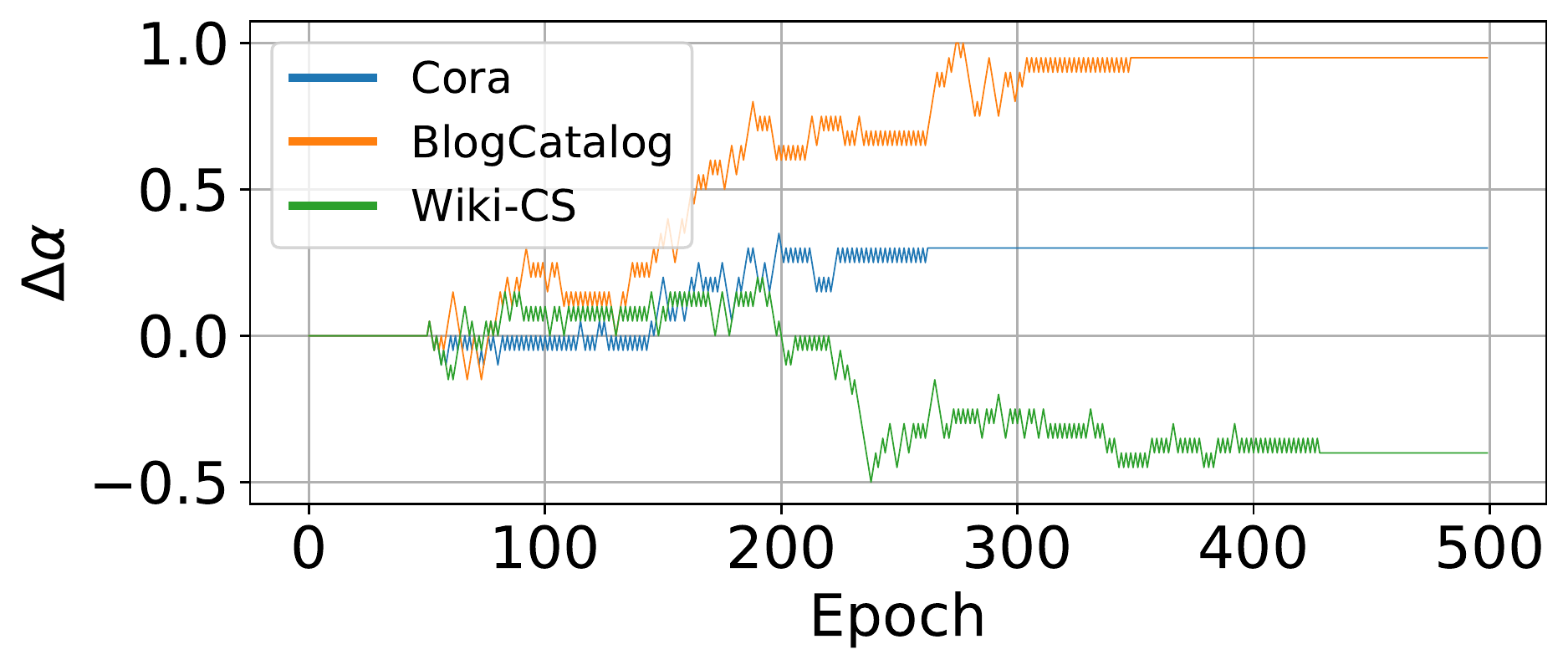}\label{fig:3b}}
	\end{center}
	\vspace{-1em}
	\caption{Reinforcement mixup mechanism analysis.}
	\vspace{-1em}
	\label{fig:3}
\end{figure}

To verify the importance of the reinforcement mixup mechanism, we remove it from GraphMixup to obtain a new variant - GraphMixup-Fix, which sets a \textit{fixed} upsampling scale for all minority classes. Then, we plot the performance curve of GraphMixup-Fix and four baselines under different (fixed) upsampling scales on the Cora dataset. As shown in Fig.~\ref{fig:3a}, we find that generating more samples for minority classes helps achieve better performance when the upsampling scale is smaller than 0.8 (or 1.0). However, when the upsampling scale becomes larger, keep increasing it may result in the opposite effect, as too many new synthesis nodes will only introduce redundant and noisy information.

Since the RL algorithm is trained jointly with GNNs, its updating and convergence process is very important. In Fig.~\ref{fig:3b}, we visualize the updating process of the cumulative change in upsampling ratio $\alpha$, e.g., $\Delta\alpha=\alpha_i-\alpha_i^{init}$. Since other modules in the framework are updated together with the RL module, the RL environment is not very stable at the beginning, so the RL algorithm starts to run only after the first 50 epochs. When the framework gradually converges, $\Delta\alpha$ bumps for several rounds and meets the terminal condition. From Fig.~\ref{fig:3b}, we find that $\Delta\alpha$ eventually converges to 0.3 on the Cora dataset, resulting in an upsampling scale $\alpha_i=\Delta\alpha+\alpha_i^{init}=0.8$ with initial value $\alpha_i^{init}=round(\frac{N}{m|C_i|})=0.5$. This corresponds to the result in Fig.~\ref{fig:3a} where $\text{GraphMixup}_C$ obtains the best performance when the upsampling scale is 0.8, which demonstrates the effectiveness of the reinforcement mixup mechanism, i.e., it adaptively determines suitable upsampling scale without the need for heuristic estimation like Fig.~\ref{fig:3a}.

\vspace{-0.5em}
\subsection{Self-Supervised Prediction Analysis (\textit{Q5})}
\begin{figure}[!htbp]
	\begin{center}
		\includegraphics[width=1.0\linewidth]{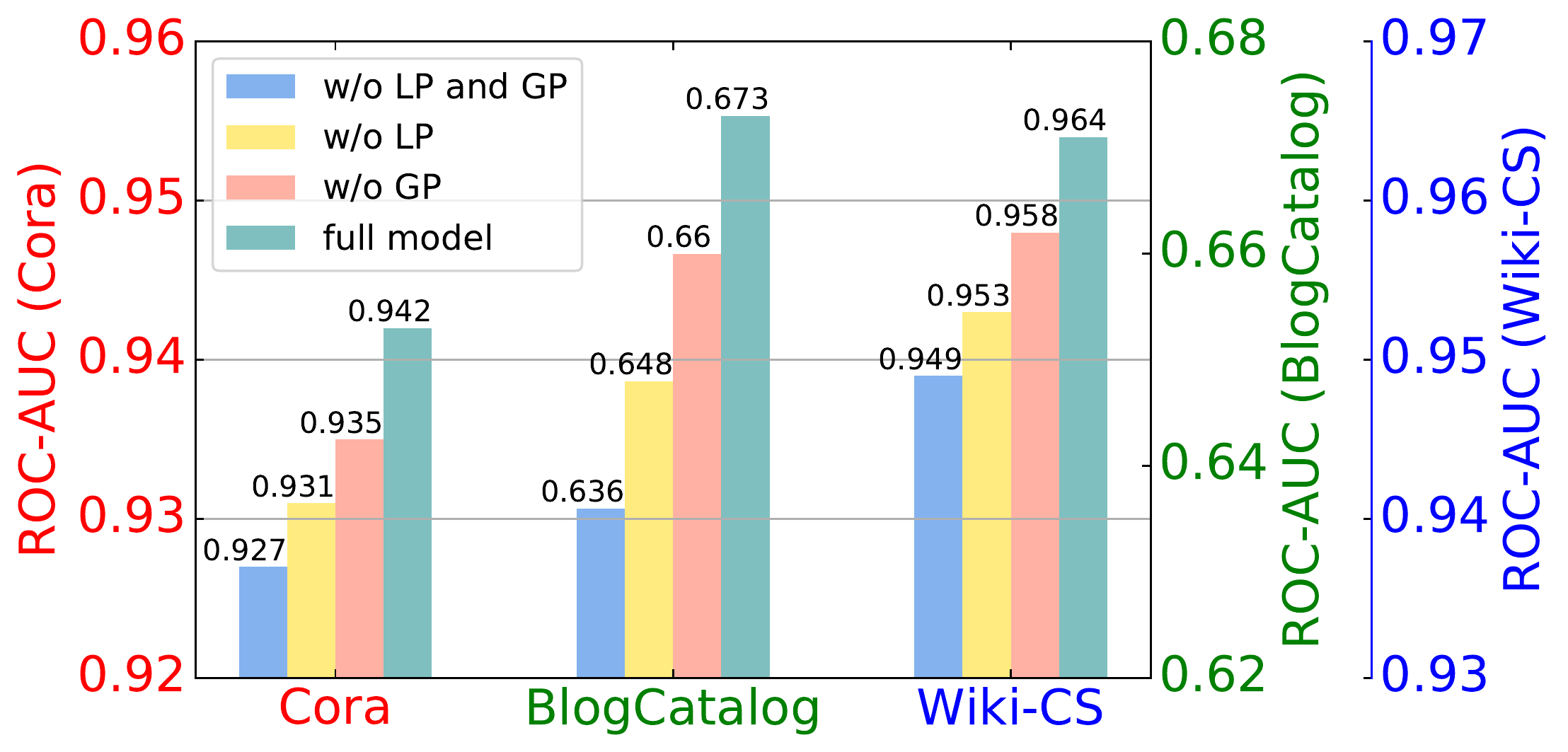}
	\end{center}
	\caption{Ablation study with different self-supervised tasks.}
	\vspace{-1em}
	\label{fig:4}
\end{figure}

This evaluates the effectiveness of self-supervised prediction tasks in the proposed framework through four sets of experiments: the model without (A) Local-Path Prediction (\emph{w/o LP}); (B) Glocal-Path Prediction (\emph{w/o GP}); (C) both Local-Path and Global-Path Prediction (\emph{w/o LP and GP}, and (D) the full model. Experiments are conducted on the Cora dataset, and ROC-AUC scores are reported as performance evaluation. After analyzing the reported results in Fig.~\ref{fig:4}, we can find that both Local-Path Prediction and Glocal-Path Prediction contribute to improving model performance. More importantly, applying these two tasks together can further improve performance on top of each of them, resulting in the best performance, which demonstrates the benefit of self-supervised prediction tasks on capturing local and global information embedded in the graph structure.

\section{Conclusion}
In this paper, we propose GraphMixup, a novel framework for improving class-imbalanced node classification on graphs. GraphMixup implements feature, label, and edge mixup simultaneously in a unified framework in an end-to-end manner. Specifically, GraphMixup performs semantic-level feature mixup by constructing semantic relation spaces and edge mixup with an edge predictor trained on two well-designed context-based self-supervised tasks; Moreover, a \emph{Reinforcement Mixup} mechanism is applied to adaptively determine the number of samples to be generated (upsampling scale) by mixup for minority classes. Extensive experiments on three real-world datasets have shown that the proposed GraphMixup outperforms other leading methods on class-imbalanced node classification tasks.

\clearpage
\bibliography{ref}

\end{document}